\documentclass{article}

\usepackage[final]{neurips_2019}

\usepackage[utf8]{inputenc}
\usepackage[T1]{fontenc}
\usepackage{hyperref}

\hypersetup{
    colorlinks=true,    
    citecolor=green,    
}

\usepackage{url}
\usepackage{booktabs}
\usepackage{amsfonts}
\usepackage{nicefrac}
\usepackage{microtype}
\usepackage{graphicx}
\usepackage{xcolor}
\usepackage{lipsum}
\usepackage{subcaption}

\usepackage{pgfplots}
\usepgfplotslibrary{groupplots}
\pgfplotsset{width=\textwidth/2,compat=newest}

\usepackage{multirow}
\usepackage{multicol}
\usepackage{array}
\newcolumntype{H}{>{\setbox0=\hbox\bgroup}c<{\egroup}@{}}
\makeatletter
\newcommand*\mysize{%
  \@setfontsize\mysize{7.5}{9.0}%
}
\makeatother

\iffalse
    
\else
    
\fi

\title{
  Are Large Language Models Useful \\ for Time Series Data Analysis?
}

\author{
Francis Tang, Ying Ding\\
School of Information, The University of Texas at Austin\\
}

\begin{document}

\maketitle

\begin{abstract}
Time series data plays a critical role across diverse domains such as healthcare, energy, and finance, where tasks like classification, anomaly detection, and forecasting are essential for informed decision-making. Recently, large language models (LLMs) have gained prominence for their ability to handle complex data and extract meaningful insights. This study investigates whether LLMs are effective for time series data analysis by comparing their performance with non-LLM-based approaches across three tasks: classification, anomaly detection, and forecasting.

Through a series of experiments using GPT4TS~\cite{GPT4TS} and autoregressive models~\cite{autotimes}, we evaluate their performance on benchmark datasets and assess their accuracy, precision, and ability to generalize. Our findings indicate that while LLM-based methods excel in specific tasks like anomaly detection, their benefits are less pronounced in others, such as forecasting, where simpler models sometimes perform comparably or better. This research highlights the role of LLMs in time series analysis and lays the groundwork for future studies to systematically explore their applications and limitations in handling temporal data.
\end{abstract}

\section{Introduction}

\subsection{Background}

Time series data, which consists of sequential measurements captured over time, is pervasive across numerous domains and applications. In healthcare, for example, time series data is used to monitor vital signs, predict patient readmissions, and track disease progression. Similarly, in the energy sector, time series data underpins forecasting of electricity demand and renewable energy generation. Weather prediction, stock market analysis, traffic management, and industrial equipment monitoring are other critical areas where time series data serves as the backbone of decision-making.

The sheer volume and complexity of time series data necessitate robust automated methods for analysis. Tasks such as \textit{classification}, where patterns within the data are identified and labeled (e.g., detecting arrhythmias from ECG signals), \textit{forecasting}, which involves predicting future values (e.g., electricity demand in the next 24 hours), and \textit{anomaly detection}, where irregular patterns are flagged (e.g., detecting cybersecurity threats in network traffic), are integral to leveraging the full potential of time series data.

Automated time series data analysis not only accelerates decision-making but also enhances accuracy, enabling data-driven insights that would otherwise require extensive manual effort. Consequently, the development of innovative methods for analyzing time series data has been an active area of research, with the advent of machine learning and, more recently, large language models (LLMs), providing new tools and opportunities for improving outcomes in this field.

\subsection{Problem Statement}

Despite the transformative potential of large language models (LLMs) in natural language processing and other domains, their applicability to time series analysis remains a topic of ongoing debate \cite{tan2024are}. Proponents of LLMs argue that their ability to process complex sequential data and capture nuanced patterns positions them as powerful tools for tasks such as classification, anomaly detection, and forecasting. Notable studies, such as those leveraging GPT-based architectures, have claimed that fine-tuned LLMs outperform traditional methods in various time series applications.

However, these claims are not without contention. Critics assert that simpler, domain-specific models, which are often computationally efficient and specifically tailored for time series data, can match or even surpass the performance of LLMs in many scenarios. For instance, models employing statistical techniques or lightweight machine learning algorithms may achieve competitive results with lower computational overhead. Furthermore, the general-purpose nature of LLMs can sometimes lead to overfitting or inefficiencies when applied to domain-specific problems, raising questions about their cost-effectiveness and scalability.

This divergence of perspectives underscores the need for a systematic evaluation of LLMs for time series analysis. By directly comparing their performance to non-LLM-based approaches across various tasks and datasets, this study seeks to provide empirical insights into their capabilities, limitations, and role within the broader ecosystem of time series analysis methods.

\subsection{Research Objective}

The primary objective of this study is to evaluate the effectiveness of large language models (LLMs) in time series data analysis by assessing their performance across three critical dimensions: \textit{classification}, \textit{anomaly detection}, and \textit{forecasting}. Specifically, the research aims to determine whether LLMs provide measurable improvements over traditional, non-LLM-based methods in these tasks.

To achieve this goal, we systematically compare the performance of LLMs, such as GPT-based architectures, with simpler, domain-specific models using a diverse set of benchmark datasets. The evaluation focuses on key metrics such as accuracy, precision, and mean squared error (MSE) to ensure a comprehensive assessment of each method's strengths and limitations. Additionally, the study explores task-specific nuances to better understand the scenarios where LLMs may excel and where simpler methods might suffice.

By addressing these questions, this research seeks to bridge the gap between existing claims and counterclaims regarding the utility of LLMs for time series data. The findings aim to inform both researchers and practitioners about the practical applicability of LLMs in solving time series challenges, contributing to the broader understanding of their role in this field.

\section{Methodology}

To evaluate the utility of large language models (LLMs) in time series analysis, we adopt a comparative framework that incorporates both LLM-based and non-LLM-based models. The framework focuses on three distinct tasks—\textit{classification}, \textit{anomaly detection}, and \textit{forecasting}—allowing for a comprehensive examination of performance across a variety of scenarios.

\begin{figure}[h]
    \centering
    \includegraphics[width=0.5\linewidth]{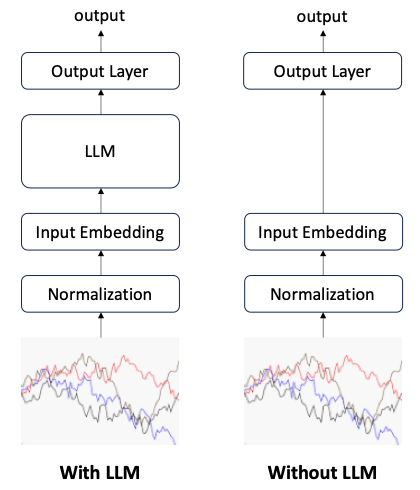}
    \caption{Time Series Model with LLM (left) and without LLM (right)}
    \label{fig:tsllm}
\end{figure}

~\autoref{fig:tsllm} presents a side-by-side comparison of two architectures for time series data analysis: one that incorporates a large language model (LLM) and another that does not. Both architectures begin with the same preprocessing pipeline. The raw time series input data, represented as overlapping line plots of sequential variables, undergoes a normalization step to standardize the data across different scales. Following normalization, the data is passed into an input embedding layer, which transforms it into a high-dimensional representation suitable for downstream processing. This stage is identical in both architectures.

The primary difference lies in the inclusion of the LLM layer in the \textit{With LLM} approach. In this architecture, the embeddings are passed through a large language model, which is specifically designed to capture complex temporal dependencies and high-level patterns in sequential data. The LLM processes the embeddings, enriching them with its ability to model long-range and non-linear relationships in the data. This enhanced representation is then passed to the output layer, which generates the final predictions or classifications.
In contrast, the \textit{Without LLM} architecture bypasses this intermediate step. Instead of leveraging the LLM’s processing capabilities, the embeddings are fed directly into the output layer, relying on simpler models or traditional statistical methods to make predictions. This approach is computationally less intensive but may not capture the same level of complexity as the LLM-enhanced architecture.

Next, we discuss two types of time series models.

\subsection{Non-Autoregressive Models}

The non-autoregressive LLM-based time series (TS) model is designed for tasks such as classification and anomaly detection, where predictions are generated for the entire input sequence simultaneously, without relying on sequential dependencies between time steps in the output.

\textbf{Input Layer:} The input to the model consists of time series data, which typically comprises multiple sequential variables observed over time. This raw input data is first normalized to ensure that all variables have a consistent scale, minimizing the effects of outliers and improving convergence during model training. The normalized data is then passed through an input embedding layer, which transforms the raw time series into high-dimensional representations. These embeddings capture both temporal and contextual features of the sequence, making them suitable for downstream processing by the LLM.

\textbf{Output Layer:} After the LLM processes the embeddings, the output layer generates predictions based on the task requirements. For classification tasks, the output layer typically applies a softmax activation function to produce probabilities for each class label. For anomaly detection, the output layer may compute a score or flag anomalies within the input sequence. Importantly, the non-autoregressive nature of this model means that the entire output is predicted simultaneously, leveraging the LLM's ability to encode complex relationships across the entire sequence.

This architecture ensures that the model captures global patterns and dependencies within the input data, making it well-suited for tasks that do not require step-by-step forecasting.

\begin{figure}[h]
    \centering
    \includegraphics[width=0.8\linewidth]{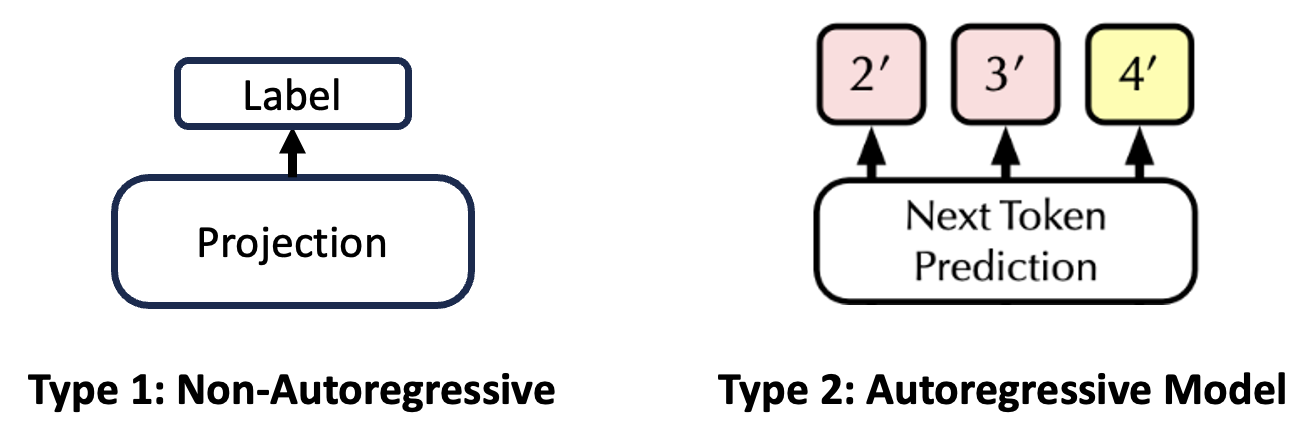}
    \caption{Left: Non-Autoregressive Model for all time series tasks \cite{GPT4TS} vs. Right: Autoregressive Model \cite{autotimes} for time series forecasting task}
    \label{fig:tsllm}
\end{figure}

\subsection{Autoregressive Models}

The autoregressive large language model (LLM)-based time series (TS) model is specifically designed for forecasting tasks, where predictions are made sequentially, with each prediction depending on prior outputs. This architecture leverages the temporal structure of time series data to provide accurate and context-aware forecasts.

\textbf{Input Layer:} The input to the autoregressive model consists of historical time series data, which represents past observations of one or more variables. Similar to non-autoregressive models, the raw input data is first normalized to ensure consistency across different features and scales. The normalized data is then passed through an input embedding layer, which encodes the temporal and contextual relationships within the sequence into a high-dimensional representation. This embedding enables the model to effectively capture patterns and dependencies from past observations, which are crucial for making future predictions.

\textbf{Output Layer:} In the autoregressive setting, the output layer generates predictions for the next time step based on both the processed input sequence and the model's prior outputs. For each predicted time step, the model updates its internal state to incorporate the new prediction before proceeding to forecast the next time step. This iterative process continues until the desired forecast horizon is reached. The output layer may include a linear transformation or activation functions (e.g., ReLU or softmax) to produce the predicted values or probability distributions, depending on the nature of the forecasting task.

The autoregressive LLM-based model is particularly well-suited for time series forecasting because of its ability to model temporal dependencies over long horizons. By sequentially incorporating prior outputs into the prediction process, the model adapts to the dynamic nature of time series data, capturing both short-term and long-term trends with high fidelity. This approach ensures that the model can handle complex forecasting tasks where future values are heavily influenced by historical patterns.

\section{Experiments}

\subsection{Classification}

We implemented GPT4TS without the LLM layer for classification tasks to evaluate its performance against the original GPT4TS model with the LLM. The experiments were conducted in an environment configured on Azure, equipped with a single NVIDIA V100 GPU and 16GB of RAM. Both models, GPT4TS without LLM and the original GPT4TS with LLM, were trained on the same datasets, ensuring a consistent evaluation framework. Hyperparameter tuning was performed for both models to optimize their performance. Finally, the best results from each model were reported, providing a direct comparison of their classification capabilities.

We evaluate the following datasets:

\begin{itemize}
    \item Readimission: 30-day short-term hospital readmission from the MIMIC-III dataset (access approved by MIT Laboratory for Computational Physiology)
    \item EthanolConcentration: water-and-ethanol solutions
    \item FaceDetection: Time series data representing pixel intensities for face detection tasks.
    \item Handwriting: Accelerometer readings capturing wrist movements during handwriting
    \item Heartbeat: Spectrogram readings of heart sounds from healthy and pathological patients.
    \item JapaneseVowels: Voice recordings of nine male speakers pronouncing Japanese vowels
    \item PEMS-SF: Traffic data from Bay Area, measuring occupancy rates across different car lanes.
    \item SelfRegulation: EEG recordings controlling a cursor’s movement via self-regulation of slow cortical potentials.
\end{itemize}

~\autoref{tab:classification} shows the results.

\begin{table}[h!]
\centering
\begin{tabular}{l|c|c}
\toprule
\textbf{Dataset} & \textbf{GPT4TS (with LLM)} & \textbf{GPT4TS (w/o LLM)} \\ \midrule
Readmission         & \textbf{94.1\%} & 93.5\% \\
EthanolConcentration & \textbf{33.1\%} & 32.3\% \\
FaceDetection        & \textbf{68.6\%} & 67.1\% \\
Handwriting          & \textbf{30.9\%} & 29.6\% \\
Heartbeat            & \textbf{76.6\%} & 74.6\% \\
JapaneseVowels       & \textbf{98.4\%} & \textbf{98.4\%} \\
PEMS-SF              & \textbf{85.0\%} & 82.1\% \\
SelfRegulation       & \textbf{92.5\%} & 91.8\% \\
\bottomrule
\end{tabular}
\caption{Comparison of accuracy between GPT4TS with LLM and without LLM on classification tasks. GPT4TS with LLM demonstrates superior performance across 7 out of the 8 datasets, achieving higher accuracy in challenging domains such as Readmission and SelfRegulation. Notably, both models achieved identical accuracy on the JapaneseVowels dataset, suggesting that the LLM layer offers minimal additional benefit for simpler tasks.}
\label{tab:classification}
\end{table}

\subsection{Anomaly Detection}

We also implemented GPT4TS with LLM and without LLM on anomaly detection task, and evaluated the follow datasets.

\begin{itemize}
    \item PSM (Pooled Server Metric): detect anomalies in server performance metrics
    \item SMD (Server Machine Dataset): detect anomalies in server operations
    \item SMAP (Soil Moisture Active Passive Satellite): detect anomalies in satellite operations
    \item SWaT (Secure Water Treatment): detect cyber-physical anomalies in water treatment processes
\end{itemize}

~\autoref{tab:anomaly_detection} shows the results.

\begin{table}[h!]
\centering
\begin{tabular}{l|c|c}
\toprule
\textbf{Dataset} & \textbf{GPT4TS (with LLM)} & \textbf{GPT4TS (w/o LLM)} \\ \midrule
PSM   & \textbf{0.971} & 0.942 \\
SMD   & \textbf{0.839} & \textbf{0.839} \\
SMAP  & \textbf{0.689} & 0.660 \\
SWaT  & \textbf{0.926} & 0.903 \\
\bottomrule
\end{tabular}
\caption{Comparison of F1-score between GPT4TS with LLM and without LLM on anomaly detection tasks. GPT4TS with LLM achieves superior F1-scores on 3 out of the 4 datasets, demonstrating significant improvements in anomaly detection for PSM, SMAP, and SWaT. Notably, both models perform equally on the SMD dataset, indicating a similar capability in detecting anomalies in server operations.}
\label{tab:anomaly_detection}
\end{table}

\subsection{Forecasting}
We implemented GPT4TS without the LLM layer for forecasting tasks to compare its performance with the original GPT4TS model and an autoregressive model. The experiments were conducted in an environment configured on LambdaLab, equipped with a single NVIDIA A6000 GPU and 48GB of RAM. Both the original GPT4TS with LLM and the GPT4TS without LLM were trained in a non-autoregressive manner, allowing for simultaneous predictions of the target sequence. Additionally, we trained the AutoTimes model, an autoregressive approach utilizing the LLaMA 7B model, for one epoch only due to the significant computational demands and long training time associated with autoregressive methods. The results from these models provide insights into the trade-offs between computational efficiency and forecasting accuracy across different architectures.

We evaluated on the follow datasets with the focus of long-term forecasting.
\begin{itemize}
    \item ETTh1 (Electricity Transformer Temperature): electricity temperature forecasting with hourly recording
    \item Weather: local climate data from 1,600 U.S. locations, between 2010 and 2013
\end{itemize}

~\autoref{tab:forecasting} shows the results.

\begin{table}[h!]
\centering
\begin{tabular}{l|c|c|c}
\toprule
\textbf{Dataset} & \textbf{GPT4TS (w/o LLM)} & \textbf{GPT4TS (with LLM)} & \textbf{AutoTimes (with LLM)} \\
\midrule
\multicolumn{4}{c}{\textbf{Evaluation Metric: MSE (Mean Squared Error)}} \\ \midrule
ETTh1   & 0.3941 & 0.3767 & \textbf{0.3582} \\
Weather & 0.1739 & 0.1659 & \textbf{0.1624} \\
\midrule
\multicolumn{4}{c}{\textbf{Evaluation Metric: MAE (Mean Absolute Error)}} \\ \midrule
ETTh1   & 0.4099 & 0.3991 & \textbf{0.3988} \\
Weather & 0.2277 & 0.2187 & \textbf{0.2135} \\
\bottomrule
\end{tabular}
\caption{Comparison of forecasting performance for GPT4TS with and without LLM, and AutoTimes with LLM (forecast length: 96). AutoTimes with LLM achieves the best performance across both datasets (ETTh1 and Weather) in terms of Mean Squared Error (MSE) and Mean Absolute Error (MAE), demonstrating the effectiveness of the autoregressive approach. GPT4TS with LLM shows improved accuracy over GPT4TS without LLM, which illustrates the usefulness of LLM even when LLM is used in non-autoregressive manner.}
\label{tab:forecasting}
\end{table}

\section{Conclusion and Future Work}

\subsection{Conclusion}

Our preliminary findings indicate that the role of large language models (LLMs) in time series analysis is significantly different from all existing work. We demonstrate that LLMs can be beneficial for various time series tasks, including classification, anomaly detection, and forecasting. However, the usage of LLMs must be tailored based on the specific task requirements:
\begin{itemize}
    \item \textbf{Non-Autoregressive Models:} Best suited for classification and anomaly detection tasks, where predictions can be generated simultaneously for the entire input sequence.
    \item \textbf{Autoregressive Models:} Essential for forecasting tasks, where predictions are made sequentially, leveraging the temporal structure of the data.
\end{itemize}

These insights provide a foundation for understanding how LLMs can be applied effectively to time series tasks, bridging the gap between theoretical capabilities and practical applications.

\subsection{Future Work}

Future studies should focus on conducting a more systematic exploration of the utility of different LLM architectures across diverse time series tasks. This includes evaluating a broader range of LLMs, datasets, and task-specific configurations to better understand their strengths and limitations. Additionally, research should investigate methods to improve the computational efficiency and scalability of LLMs in time series applications, ensuring their practicality for real-world use cases.

\bibliography{bibliography}

\begin{thebibliography}{3}
\providecommand{\natexlab}[1]{#1}
\providecommand{\url}[1]{\texttt{#1}}
\expandafter\ifx\csname urlstyle\endcsname\relax
  \providecommand{\doi}[1]{doi: #1}\else
  \providecommand{\doi}{doi: \begingroup \urlstyle{rm}\Url}\fi

\bibitem[Zhou et~al.(2023)Zhou, Niu, Wang, Sun, and Jin]{GPT4TS}
Tian Zhou, Peisong Niu, Xue Wang, Liang Sun, and Rong Jin.
\newblock One fits all: Power general time series analysis by pretrained {LM}.
\newblock In \emph{Thirty-seventh Conference on Neural Information Processing Systems}, 2023.

\bibitem[Liu et~al.(2024)Liu, Qin, Huang, Wang, and Long]{autotimes}
Yong Liu, Guo Qin, Xiangdong Huang, Jianmin Wang, and Mingsheng Long.
\newblock Autotimes: Autoregressive time series forecasters via large language models.
\newblock In \emph{The Thirty-eighth Annual Conference on Neural Information Processing Systems}, 2024.

\bibitem[Tan et~al.(2024)Tan, Merrill, Gupta, Althoff, and Hartvigsen]{tan2024are}
Mingtian Tan, Mike~A Merrill, Vinayak Gupta, Tim Althoff, and Thomas Hartvigsen.
\newblock Are language models actually useful for time series forecasting?
\newblock In \emph{The Thirty-eighth Annual Conference on Neural Information Processing Systems}, 2024.
\newblock URL \url{https://openreview.net/forum?id=DV15UbHCY1}.

\end{thebibliography}
\bibliographystyle{unsrtnat}

\end{document}